\documentclass[runningheads]{llncs}
\usepackage[T1]{fontenc}
\usepackage{graphicx}
\usepackage{colortbl}
\usepackage{xcolor}
\usepackage{amsmath,amssymb,amsfonts}
\usepackage{multirow}
\usepackage{marvosym}
\usepackage{wasysym} 
\begin{document}

\title{Gated Fusion Enhanced Multi-Scale Hierarchical Graph Convolutional Network for Stock Movement Prediction}
\titlerunning{Gated Fusion Multi-Scale Hierarchical GCN for Stock Prediction}
%
\author{Xiaosha Xue\inst{1} \and
Peibo Duan\inst{1} \textsuperscript{(\Letter)} \and
Zhipeng Liu\inst{1} \and Qi Chu\inst{2} \and Changsheng Zhang\inst{1} \and Bin Zhang\inst{1}}
\authorrunning{X. Xue et al.}
%
\institute{College of Software, Northeastern University, 110819, Shenyang, China \email{2371473@stu.neu.edu.cn}, \email{sakuragiduan@gmail.com}, \email{2310543@stu.neu.edu.cn}, \email{\{zhangchangsheng,zhangbin\}@mail.neu.edu.cn} \and
Neusoft Corporation
\email{chu\_q@neusoft.com}}
\maketitle              
\begin{abstract}

Accurately predicting stock market movements remains a formidable challenge due to the inherent volatility and complex interdependencies among stocks. Although multi-scale Graph Neural Networks (GNNs) hold potential for modeling these relationships, they frequently neglect two key points: the subtle intra-attribute patterns within each stock affecting inter-stock correlation, and the biased attention to coarse- and fine-grained features during multi-scale sampling. To overcome these challenges, we introduce MS-HGFN (Multi-Scale Hierarchical Graph Fusion Network). The model features a hierarchical GNN module that forms dynamic graphs by learning patterns from intra-attributes and features from inter-attributes over different time scales, thus comprehensively capturing spatio-temporal dependencies. Additionally, a top-down gating approach facilitates the integration of multi-scale spatio-temporal features, preserving critical coarse- and fine-grained features without too much interference. Experiments utilizing real-world datasets from U.S. and Chinese stock markets demonstrate that MS-HGFN outperforms both traditional and advanced models, yielding up to a 1.4\% improvement in prediction accuracy and enhanced stability in return simulations. The code is available at https://github.com/snowman0123/MS-HGFN.

\keywords{Trend prediction \and Graph Neural Networks \and Multi-scale Fusion.}
\end{abstract}

\section{Introduction}

The stock market is crucial to the global financial system, and precise prediction of stock trends is essential for crafting profitable investment strategies \cite{c01,c02}. This is challenging due to stock price volatility and non-stationarity. Nonetheless, advancements in deep learning \cite{c03,c34} present new opportunities for portfolio optimization.

Recent research on stock prediction focuses on two main areas. The first area involves forecasting stock trends through historical price indicators, with Transformers \cite{c38,c25} being frequently employed for their ability to handle global temporal dependencies via self-attention. The second area focuses on concurrently capturing both temporal and spatial features, where spatial features are defined by the interactions between stocks, reflecting the complex interactive network of the stock market\cite{c39,c40}. To model these interactions between stocks, Graph Neural Networks (GNNs) \cite{c04} have been employed, which treat stocks as nodes and their relationships as edges to capture relational (spatial) dependencies. Moreover, multi-scale analysis approaches have been proven to be effective tools for solving trend prediction problems, as they can capture more temporal patterns in stock data at different time scales \cite{c08}. For example, daily stock price indicators exhibit the fine-grained fluctuations of stocks, while weekly and monthly data reflect the coarser-grained and long-term trends \cite{c30}. Therefore, exploring multi-scale GNN frameworks is considered a promising way to capture the diverse spatial dependencies of stocks at different scales \cite{c31}. 

\begin{figure*}[ht]
    \centering
    \includegraphics[width=\textwidth]{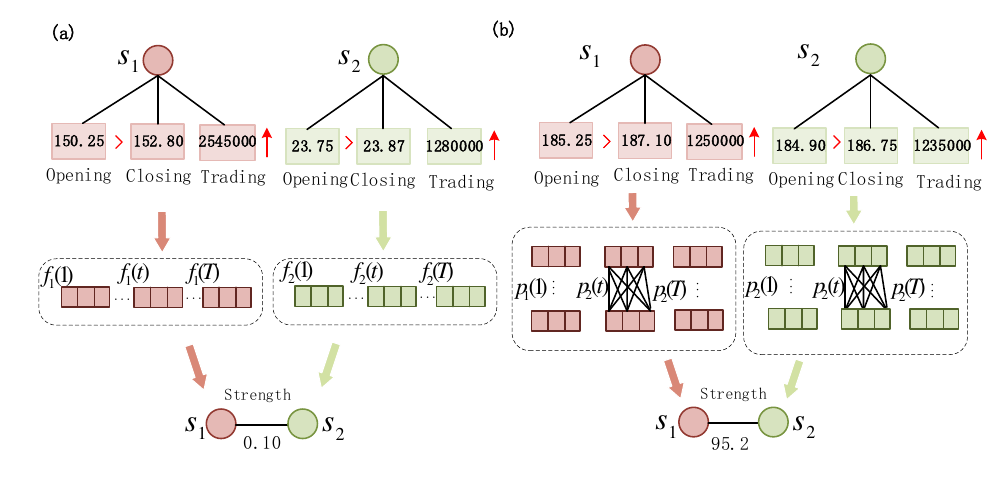}
    \caption{The schematic diagram of correlation between two stocks by considering inter-attributes and intra-attributes}
    \label{fig:sample-image}
\end{figure*}

However, a majority of multi-scale GNN methods exhibit two significant limitations. First, the implicit relational dependencies between stocks, especially the strength of the dependencies, are mainly inferred from the correlation between feature embeddings extracted based on all attributes pertaining to each stock, e.g., the feature embedding of closing and opening prices of a stock. It disregards the pattern hidden in the intrinsic interactions within a stock's attributes that may correlate with those in other stocks, affecting inter-stock relational dependencies. As shown in Fig.1 where attributes such as opening price, closing price, and trading volume are considered for stock $S_1$ and $S_2$. With conventional methods (Fig.1(a)), two feature embeddings, $f_1(t)$ and $f_2(t)$, are extracted from the time series of three attributes for $S_1$ and $S_2$. There is an edge with weak correlation (e.g., 0.10) between $S_1$ and $S_2$ due to the significant disparity between $f_1(t)$ and $f_2(t)$. This is attributed to the fact that $S_1$ exhibits high values across all three attributes contrasted with the low values observed in $S_2$. In Fig.1(b), an edge with strong correlation (e.g., 95.2) is established between $S_1$ and $S_2$ because both stocks demonstrate analogous intrinsic patterns ($p_1(t)$ and $p_2(t)$) within these attributes. Specifically, both stocks show a rise in the closing price that notably exceeds the opening price, coupled with a significant increase in trading volume. This analogous pattern indicates that the price trend variations of the two stocks are indeed similar. 

Secondly, existing multi-scale fusion techniques tend to diminish the role of coarse-grained features critical for stock trend prediction. In stock trend prediction, fine-grained time scales intend to capture microscopic information like local patterns in price fluctuations, while coarse-grained scales are more helpful in revealing macroscopic information like trends in the market \cite{c21}. Current methods predominantly use a weighted approach, like attention mechanisms, to merge multi-scale features for trend forecasting \cite{c30}. In practice, as scale decreases, detailed information becomes prevalent, reducing the impact of coarse-grained trend information crucial for prediction accuracy. Additionally, simple fusion methods can lead to feature misalignment and redundancy, hindering the optimal use of multi-scale information and thus affecting prediction accuracy.

In response to the aforementioned limitations, we introduce an innovative framework termed the Multi-Scale Hierarchical Graph Fusion Network (MS-HGFN). Two core modules are designed in this framework, namely the hierarchical Graph Neural Network (GNN) module and the gated fusion module. More precisely, the hierarchical GNN module integrates learnable matrices to capture the pattern in terms of the intrinsic interactions within a stock's attributes at each time scale. The gated fusion module utilizes a top-down strategy with a gating mechanism to maintain essential coarse- and fine-grained features at small time scales. The contributions can be summarized as follows:

 \begin{itemize}
        \item The proposed hierarchical GNN module enhances the detection of dynamic spatio-temporal dependencies between stocks by analyzing features across both inter-attribute and intra-attribute dimensions.       
        \item The novel gating-based multi-scale fusion approach balances coarse- and fine-grained features to further enhance prediction accuracy. 
        \item Extensive experiments on three authentic stock datasets from the US and China markets reveal our method surpasses state-of-the-art methods in accuracy, achieving up to a $1.4\%$ enhancement.
    \end{itemize}

\section{Related Work}

\subsection{Stock time series prediction models}

Due to the temporal nature of stock prices, researchers typically employ time series models, such as RNNs and their variants, including LSTMs and GRUs, for their ability to capture sequential dependencies. For instance, Nelson et al. \cite{c06} utilized LSTM to predict future stock price movements by leveraging historical price indicators, demonstrating the effectiveness of LSTM in capturing temporal dependencies in financial time series. Smith et al. \cite{c15} introduced an optimized LSTM-RNN model designed to improve investors' capabilities in navigating dynamic stock market environments. This advancement enables more accurate predictions of stock trends.

Researchers have found that combining LSTMs and GRUs with NLP and text mining improves stock price prediction by incorporating macroeconomic conditions, investor sentiment, and company-specific news. Zhuge \cite{c10} utilized an LSTM combined with sentiment analysis via a naive Bayesian classifier to capture non-linear and long-term dependencies in stock price predictions. Chen et al. \cite{c16} improved predictions by integrating quantitative indicators and sentiment-based news features into their RNN-boost model. Gong et al. \cite{c17} introduced MSHAN, which enhanced daily stock price direction predictions on the StockNet dataset by merging historical data with social media insights, potentially boosting trading returns.

Unlike traditional RNN models, Transformer \cite{c33} uses a self-attention mechanism to manage long-distance dependencies and parallelize computations, enhancing training efficiency. Zhang et al. \cite{c32} introduced TEANet, which leverages a Transformer encoder to integrate social media and stock prices, boosting prediction accuracy. Muhammad et al. \cite{c35} enhanced stock price prediction by combining Transformer with time2vec encoding for better predictive capabilities.

\subsection{Graph Neural Network Models}

In recent years, a novel research direction has focused on investigating graph-structured data to capture the interconnections between stocks \cite{c08,c12}. This approach can be categorized into two types: explicit-based and implicit-based methods. Cheng R et al. \cite{c20} developed AD-GAT, modeling momentum spillovers by integrating firm attributes with market dynamics to enhance predictions. Gao J et al. \cite{c22} proposed TRAN, which uses stock relation graphs and time-varying correlation strengths for effective stock recommendation by ranking stocks based on return ratios.

While explicit relationships may reveal superficial connections, graphs constructed using these methods are often limited by fixed, predefined corporate relationships that depend heavily on established external knowledge. Consequently, capturing deeper interconnections has proven challenging. To address this, techniques have emerged to uncover implicit relationships \cite{c13,c14}. You Z et al. \cite{c19} introduced DGDNN, which improves stock movement prediction by constructing dynamic stock graphs and learning inter-stock dependencies through graph diffusion and decoupled representation. Du K et al. \cite{c23} developed a dynamic dual-graph neural network that integrates price and semantic relationships via graph attention to capture complex interrelations. Liu et al. \cite{c31} combined multi-time scale learning with multi-graph attention networks, boosting stock index prediction performance. Although graph neural network stock prediction models based on implicit relationships have made some progress, these methods mainly focus on identifying correlations between stocks and often overlook the correlations between internal features of stocks.


\begin{figure*}[ht]
    \centering
    \includegraphics[width=\textwidth]{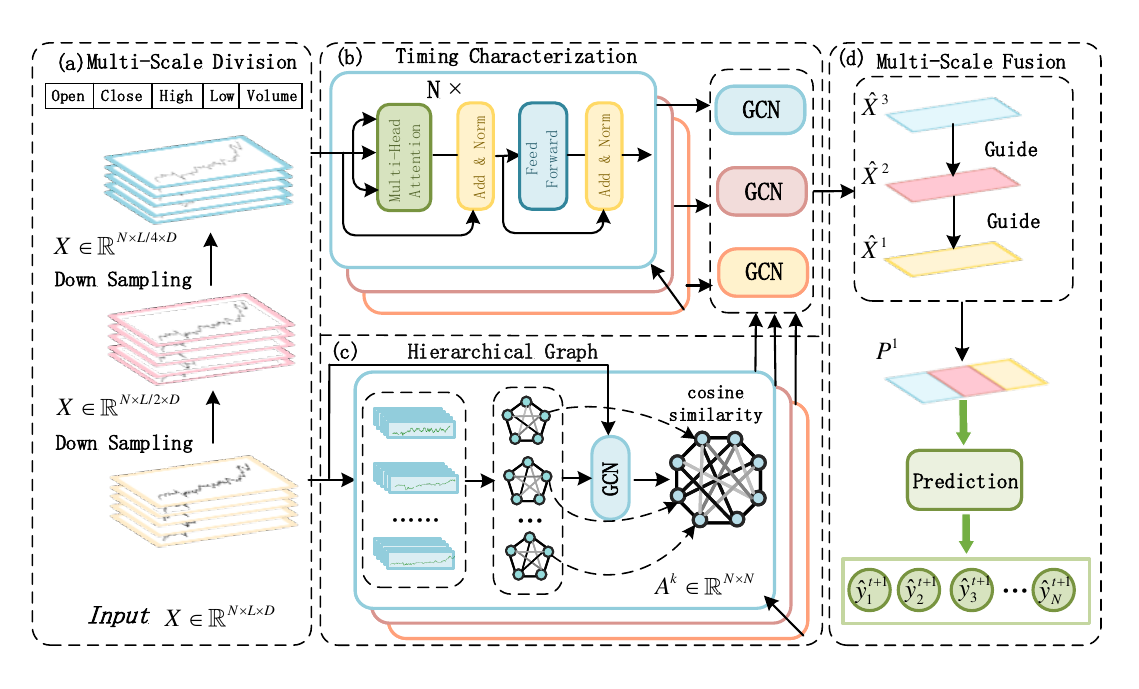}
    \caption{Overall structure of MS-HGFN. (a) Average pooling is used for K-scale sampling. (b) Temporal dependencies are explored using transformer backbone. (c) Adjacent matrices are created at each scale to dynamically model spatial dependencies using patterns learned from intra-attributes and features learned from inter-attributes. (d) Top-down multi-scale feature fusion is performed at different scales to balance coarse- and fine-grained features. Finally, the fused features are sent to the prediction module.}
    \label{fig:sample-image}
\end{figure*}

\section{Methodology}

\subsection{Problem Formulation}

We formulate stock trend prediction as a binary node classification task. Let $V = \{v_1,v_2,...,v_N\}$ denote the set of $N$ stocks. The historical price indicator for $\forall v_i \in V$ is represented as $X_i=\{{x}_{i}^{t-L+1},{x}_{i}^{t-L+2},...,{x}_{i}^{t}\}\in {\mathbb{R}}^{L\times D}$, where $D$ is the number of price indicators. A dynamic stock graph on trading day $t$ is defined as ${G}_{t}=\left \{ V, X, {A}_{t}\right \}$, where $A_t \in \mathbb{R}^{N \times N}$ denotes the weighted stock adjacency matrix constructed by analyzing the correlations among stock price indicators. We aim to develop a multi-scale GNN framework, i.e., MS-HGFN, for stock trend prediction. Mathematically, the problem can be expressed as:

\begin{equation}  
    \hat{y}^{t+1} = \text{MS-HGFN}(X;\Theta),
\end{equation}

\noindent where $X\in {\mathbb{R}}^{N \times L\times D}$ serves as the input for the model, representing the $D$ price indicators, including opening, closing, highest, lowest prices, and trading volume, for $N$ stocks across a sequential window of $L$ trading days, and $\Theta$ denotes the learnable parameter set of the model. The predicted probability at time $t+1$ is represented as $\hat{y}^{t+1}=\{\hat{y}^{t+1}_{1},\hat{y}^{t+1}_{2},...,\hat{y}^{t+1}_{n}\}$.  For $\forall v_i\in V$, the movement trend label ${y}^{t+1}_{i}$ can be formulated as:

\begin{equation}
   {y}^{t+1}_{i}=\left\{\begin{matrix}
 1,& (p_i^{t+1}-p_i^{t})/p_i^{t} \ge \gamma,\\ 
 0,& otherwise.
\end{matrix}\right.
\end{equation}

\noindent where $p_i^{t+1}$ represents the closing price of $v_i$ at time $t+1$, and the value of $\gamma$ is 0.005.
The objective of this model is to predict whether each stock will grow or fall at the next time step $t+1$. For each stock, the model generates a prediction probability, representing the probability that the stock will rise in the next time step. For this purpose, the forthcoming section will provide an in-depth account of the model, encompassing the multi-scale sampling module, the dynamic hierarchical relationship modeling module, and the multi-scale feature fusion module.

\subsection{Multi-scale Sampling Module}

To explore the potential of multiscale time series in time variation modeling and prediction tasks, we first downsample the time series $X\in {\mathbb{R}}^{N \times L\times D}$ to $K$ scales using an average pooling:

\begin{equation}
    {X}^{k} = \text{AvgPool}_{2^{k-1}}({X}^{k-1}), \quad k\in \{1,...,K\},
\end{equation}

\noindent where ${X}^{k} \in \mathbb{R}^{N \times\lceil L/2^{k-1} \rceil \times D}$ represents the downsampled series at the $k$ time scale and $K$ is the total layer. Each scale-specific time series is treated as an independent input for the subsequent module.

\subsection{Hierarchical GNN Module}

\subsubsection{Temporal Dependencies}

Inspired by the capacity of a Transformer to effectively capture extended temporal dependencies, we utilize an encoder-only framework to convert the input historical data into high-dimensional representations. Specifically, time series of each scale are mapped to the latent space of dimension $D$ via a trainable linear projection ${W}^{k}\in {R} ^{D\times D}$. The input fed into the Transformer encoder is represented as ${X}_{d}^{k}={X}^{k}{W}^{k}$, where ${X}_{d}^{k}\in \mathbb{R}^{N\times L_k\times D}$ and $L_k=\left \lceil L/2^{k-1} \right \rceil$. For each attention head $h\in \{1,2,...,H\}$, the input is projected into query, key, and value matrices:

\begin{equation}
Q^{k}_h = (X^{k}_d) W^Q_h, K^{k}_h = (X^{k}_d) W^K_h,  V^{k}_h = (X^{k}_d) W^V_h,
\end{equation}

\noindent where $W^Q_h,W^K_h,W^V_h\in \mathbb{R}^{D\times d_h}$ are learnable parameters and $ d_h=D/H$. These transformations adaptively capture dependencies specific to scale $k$. The attention output is calculated using scaled dot-product attention:

\begin{equation}
    O^{k}_h = \text{Softmax}\left(\frac{Q^{k}_h (K^{k}_h)^T}{\sqrt{d_h}}\right) V^{k}_h,
\end{equation}

\noindent where $O^{k}_h \in\mathbb{R}^{N\times L_k \times d_h}$. Finally, the outputs from multi-head attention modules are concatenated into the linear layer, followed by a residual connection and layer normalization:

\begin{equation}
    Z^{k} = \text{LayerNorm}({X}_{d}^{k}+Concat( O^{k}_1,..., O^{k}_H)W^O),
\end{equation}

\noindent where $W^O\in \mathbb{R}^{(H\times d_h)\times D}$ projects the concatenated features. The final output of the Transformer encoder is denoted as  $Z^{k} \in \mathbb{R}^{N\times D}$, capturing the temporal dependencies in the time series at scale $k$.

\subsubsection{Spatial Dependencies}

To adequately capture spatial dependencies in stock time series, the proposed hierarchical GNN module employs two types of graphs that illustrate spatio-temporal dependencies among stocks by examining features across inter-attribute dimensions and patterns within intra-attribute dimensions. First, we calculate the attribute relationships of each stock. For the $i$-th stock, we randomly initialize two adaptive matrices, ${E}_{1}^{k}$ and ${E}_{2}^{k}\in \mathbb{R}^{D\times h}$, which are optimized during the training phase. We use matrix multiplication to generate adaptive representations for capturing the correlations between attributes, which is expressed as:
\begin{equation}
    {R}_{i}^{k} = \text{Softmax}\left(\text{ReLU}\left({E}_1^{k}({E}_2^{k})^\top\right)\right).
\end{equation}

After obtaining the attribute adjacency matrix of $N$ stocks on the $k$-th scale, we use GCN to capture the correlation between stocks. The formula is as follows:

\begin{equation}
    e_{i}^{k} = \text{GCN}(\tilde{R}_{i}^k, X_{i}^k, W_{\text{c}}^{k}),
\end{equation}

\noindent where $\tilde{R}_{i}^k={R}_{i}^{k}+I$ is used to preserve original features, $X_{i}^k$ denotes the time data of the $i$-th stock at the $k$-th scale, ${e}_{i}^{k}\in \mathbb{R}^{D\times D}$ represents the node embeddings obtained from stock attributes after applying GCN. To capture the relationship between stocks, we perform a flattening operation on ${e}_{i}^{k}$ to obtain vector $\tilde{e}_{i}^{k}\in \mathbb{R}^{D^2}$. Subsequently, cosine similarity is used to ascertain the inter-node correlation, leading to the derivation of the global adjacency matrix, as articulated below:

\begin{equation}
    {A}^{k}=\frac{\tilde{e}_{i}^{k} \cdot \tilde{e}_{j}^{k}}{|\tilde{e}_{i}^{k}| |\tilde{e}_{j}^{k}|},
\end{equation}

\noindent where ${A}^{k}\in \mathbb{R}^{N\times N}$ represents the complex relationships between stocks extracted through attribute relationships, which are then served as independent inputs to subsequent processing modules.

\subsubsection{Spatio-temporal Feature Embedding}

To combine spatial-temporal features from different time scales, we use GCN to process the temporal features $Z^{k}$ obtained from the Transformer encoder and the stock relationships ${A}^{k}$ captured by the hierarchical graph module, which helps effectively integrate spatial-temporal data features at the $k$-th scale. The fusion process is mathematically expressed as:

\begin{equation}
    \hat{{X}}^{k} = \text{GCN}\left({A}^{k},{Z}^{k},{W}_g^{k} \right),
\end{equation}

\noindent where ${W}_g^{k}\in {\mathbb{R}}^{D\times D}$ denotes the trainable parameters of GCN in the $k$ scale, $ \hat{{X}}^{k}\in {\mathbb{R}}^{N\times D}$. Through this process, GCN captures the spatial relationships and temporal dependencies of stocks, which can provide more complete information for subsequent modules to use.

\subsection{Multi-scale Feature Fusion Module}

We propose a novel method for multi-scale information fusion, where rich trend features from the top gradually guide fine-grained features at the bottom. Firstly, a linear transformation is applied to each scale to map all scales to the same dimension.

\begin{equation}
P^{k+1}=\hat{{X}}^{k}W^{k}_{l} \quad k\in \{1,...,K\},
\end{equation}

\noindent where $W^{k}_{l}\in {\mathbb{R}}^{D\times D}$ is a learnable weight matrix designed to project features into a unified dimensional space. This conversion ensures size compatibility between different scales. In order to better utilize the rich trend features at the top level to guide the fine-grained features at the bottom level, we introduce a gating mechanism. The following equation illustrates how to use a gating mechanism to filter trend information:

\begin{equation}
\begin{aligned}
& \alpha ^{k}=\sigma ([\hat{{X}}^{k}||P^{k+1}]W_a), \\
& P^{k} =  \text{LayerNorm}(\alpha ^{k}\odot \hat{{X}}^{k}+(1-\alpha ^{k})\odot P^{k+1}), \\
&
\end{aligned} 
\end{equation}
\noindent where $W_a\in {\mathbb{R}}^{2D\times 1}$ is a learnable matrix, and $\sigma$ representing the sigmoid function, there $\alpha ^{k} \in (0,1)$ is a gating vector that can adjust the weights of information from different scales to preserve important trend information while reducing the impact of noise. Here, $\odot$ is multiplied element by element. Each scale is fused with the previous scale, and $P^{1}$ is the fused final scale which integrates multi-scale temporal dependencies.

\subsection{Prediction}

Finally, we utilize $P^{1}$ obtained from multi-scale fusion and use linear transformation and prediction functions for prediction,  which is formally represented as:

\begin{equation}
    \hat{{y}}^{t+1}_{i} = \text{Prediction}\left(P^{1}{W}_p+{b}_p\right),
\end{equation}

\noindent where ${W}_p\in {\mathbb{R}}^{D\times C}$ and bias ${b}_p\in {\mathbb{R}}^{C}$ are learnable parameters. The $C=2$ represents the number of classes. $Prediction(·)$ is a feedforward network with two fully connected layers.

\begin{table}[ht]
    \caption{\textnormal{Setting of Dataset}}
    \scriptsize
    \setlength{\arrayrulewidth}{0.7pt}
    \renewcommand{\arraystretch}{1.5}
    \centering
    \setlength{\tabcolsep}{6pt}
    \begin{tabular}{>{\hspace{8pt}}c c c c }
    \hline
    \multicolumn{1}{>{\hspace{8pt}}c}{\cellcolor[rgb]{0.867, 0.875, 0.91}} & 
    \multicolumn{1}{c}{\cellcolor[rgb]{0.867, 0.875, 0.91}S\&P100} &
    \multicolumn{1}{c}{\cellcolor[rgb]{0.867, 0.875, 0.91}CSI300} & 
    \multicolumn{1}{c}{\cellcolor[rgb]{0.867, 0.875, 0.91}CSI500}\\
    
    \hline
    \rowcolor[rgb]{.949, .949, .949}
    \multirow{2}{*}[0.5em]{\shortstack{Stocks}}& 96 & 196 & 325 \\
    \multirow{2}{*}[0.5em]{Periods} & 2019.1.1-2023.9.30 & 2019.1.1-2023.9.30 & 2019.1.1-2023.9.30 \\

    \rowcolor[rgb]{.949, .949, .949}
    \multirow{2}{*}[0.5em]{Trading Days} & 1195  & 1154 & 1154 \\
    \multirow{2}{*}[0.5em]{Train:Val:Test} & 896:149:150   & 866:144:144 & 866:144:144 \\
    \hline
    \end{tabular}
    \label{tab:my_label}
\end{table}

\section{Experiments}

\subsection{Dataset and Experimental Setting}

\subsubsection{Dataset}To validate our stock trend prediction method, we conduct experiments using three benchmark datasets: S\&P 100 (U.S. market), CSI 300 and CSI 500 (China market). The data, from January 1, 2019, to September 30, 2023, is split chronologically: 75\% for training, 12.5\% for hyperparameter tuning, and 12.5\% for evaluation. Stocks with incomplete records due to suspensions or delistings are excluded. Five daily price indicators (opening, closing, highest, lowest prices, trading volume) are used as features and normalized with Z-score before model input. Table 1 shows detailed statistics.
For model training, we apply a 0.5 Dropout rate at each layer to prevent overfitting. MS-HGFN parameters are trained with Adam optimizer on an NVIDIA GeForce RTX 4090 GPU for 50 epochs, with a learning rate of 1e-4 and a batch size of 32.

\subsubsection{Experimental Setting} We consider trend prediction as a binary classification problem. If the closing price is higher than the opening price, the sample is marked as "up"; otherwise, it is marked as "down". To assess the performance of the method, we conduct the experiments with six trend prediction methods for comparison, i.e., GCN \cite{c04}, TGC \cite{c37}, VGNN \cite{c29},ADGAT \cite{c20},MDGNN \cite{c28},Transformer \cite{c33}. We use accuracy (ACC) and Matthews correlation coefficient (MCC) as evaluation metrics.

\begin{table}[ht]
    \caption{\textnormal{Performance comparison with baseline models on S\&P100, CSI300 and CSI500. Bold and underlined text represent the best and second-best results, respectively.}}
    \scriptsize
    \setlength{\arrayrulewidth}{0.7pt}
    \renewcommand{\arraystretch}{1.5}
    \centering
    \setlength{\tabcolsep}{4pt}
    \begin{tabular}{>{\hspace{8pt}}c c c c c c c c c}
    \hline
    \multicolumn{1}{>{\hspace{8pt}}c}{\cellcolor[rgb]{0.867, 0.875, 0.91}Dataset} & 
    \multicolumn{1}{c}{\cellcolor[rgb]{0.867, 0.875, 0.91}Metrics} &
    \multicolumn{1}{c}{\cellcolor[rgb]{0.867, 0.875, 0.91}GCN} & 
    \multicolumn{1}{c}{\cellcolor[rgb]{0.867, 0.875, 0.91}TGC} & 
    \multicolumn{1}{c}{\cellcolor[rgb]{0.867, 0.875, 0.91}VGNN} & 
    \multicolumn{1}{c}{\cellcolor[rgb]{0.867, 0.875, 0.91}ADGAT} & 
    \multicolumn{1}{c}{\cellcolor[rgb]{0.867, 0.875, 0.91}MDGNN} & 
    \multicolumn{1}{c}{\cellcolor[rgb]{0.867, 0.875, 0.91}Transformer} & 
    \multicolumn{1}{c}{\cellcolor[rgb]{0.867, 0.875, 0.91}MS-HGFN} \\
    
    \hline
    \rowcolor[rgb]{.949, .949, .949}
    \multirow{2}{*}[0.5em]{\shortstack{S\&P100}} & ACC & \underline{52.92} & 51.89 & 50.93 & 51.01 & 48.15 & 51.29 & \textbf{53.35} \\
    \rowcolor[rgb]{.949, .949, .949}
     & MCC & \underline{0.0447} & 0.0357 & 0.0482 & 0.0194 & 0.0109 & 0.0590 & \textbf{0.0612} \\
    \hline
    
    \multirow{2}{*}[0.5em]{CSI300} & ACC & \underline{52.45} & 51.80 & 50.06 & 50.18 & 51.58 & 50.19 & \textbf{53.21} \\
     & MCC & \underline{0.0333} & 0.0292 & 0.0406 & 0.0133 & 0.0262 & 0.0359 & \textbf{0.0567} \\
    \hline
    
    \rowcolor[rgb]{.949, .949, .949}
    \multirow{2}{*}[0.5em]{CSI500} & ACC & \underline{52.06} & 51.65 & 50.11 & 50.01 & 48.09 & 50.04 & \textbf{53.46} \\
    \rowcolor[rgb]{.949, .949, .949}
     & MCC & \underline{0.0297} & 0.0192 & 0.0266 & 0.0127 & 0.010 & 0.294 & \textbf{0.0651} \\
    \hline
    \end{tabular}
    \label{tab:my_label}
\end{table}

\subsection{Evaluation Metrics}

In classification problems, ACC and MCC are often used as evaluation metrics \cite{c36}. ACC is used to provide overall performance, while MCC provides feedback on the model's performance in considering data imbalance and classification quality. Therefore, we use ACC and MCC to evaluate the stock market trend prediction ability of all methods. The higher the values of these two indicators, the better the model performance.

\begin{figure*}[ht]
    \centering
    \scalebox{1}{\includegraphics[width=\textwidth]{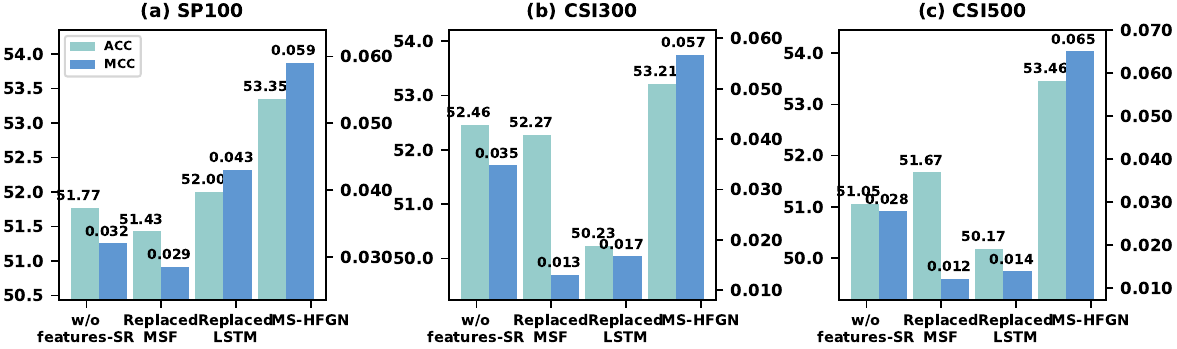}}
    \caption{Ablation study}
    \label{fig:sample-image}
\end{figure*}

\subsection{Baselines}
To evaluate the performance of the proposed MS-HGFN, we compared it with various classic and state-of-the-art methods, and all baselines are briefly described as follows:
\begin{itemize}
    \item \textbf{GCN} (2016) \cite{c04}: GCN predicts stocks by aggregating pre-defined relationships of stocks. We compare a model structure that includes an LSTM layer and two GNN layers.  
    \item \textbf{Transformer} (2017) \cite{c33}: Transformer has advantages in capturing long-range dependencies in sequence data. We adopted a Transformer model with specified hyperparameters to predict stock price trends, which is consistent with the stock price indicators used in our main model MS-HGFN to ensure data input consistency between models.
    \item \textbf{TGC} (2019) \cite{c37}:  Time Graph Convolution (TGC) can learn implicit stock relationships, combine GNN with temporal information, and integrate LSTM for stock ranking prediction.
    \item \textbf{ADGAT} (2021) \cite{c20}: Attribute-Driven Graph Attention Networks(ADGAT) used attribute sensitive momentum overflow modeling and unshielded attention mechanism, utilizing a new tensor based feature extractor to dynamically infer company relationships from market signals to improve performance.

    \item \textbf{VGNN} (2023) \cite{c29}:  Vague Graph (VGNN) demonstrates significant performance improvements in fuzzy graph learning tasks using a novel decoupled graph learning framework.
    \item \textbf{MDGNN} (2024) \cite{c28}:  Multi relational dynamic graph neural network (MDGNN) combines multiple types of relationships and temporal dynamics to model complex, evolving relationships in multi relational graph data.

\end{itemize}

\subsection{Performance Comparison}

We compared the MS-HGFN model with several baseline methods for predicting stock trends, yielding notable findings as outlined in Table 2.

MS-HGFN demonstrated superior performance across all datasets in terms of accuracy (ACC) and Matthews correlation coefficient (MCC). For instance, in the S\&P100 dataset, the model attained the highest ACC of 53.35\%, outperforming the next best model, GCN, by 0.43\%. It also achieved the highest MCC of 0.0590. Similarly, in the CSI300 dataset, MS-HGFN achieved an ACC of 53.21\%, surpassing all baselines, with GCN as the closest competitor at 52.45\%. The MCC reached 0.0567, further highlighting its superior performance. In the CSI500 dataset, the model maintained its lead with an ACC of 53.46\% and an MCC of 0.0651, outperforming models like GCN and VGNN.

These results underscore the MS-HGFN model's effectiveness in capturing complex inter-stock relationships and temporal dependencies, which enhances prediction accuracy across various datasets. The model's ability to integrate stock relationships and time dependencies provides a comprehensive understanding of market dynamics, making it a robust tool for predicting stock market trends. By modeling implicit relationships, MS-HGFN improves predictive performance, essential for navigating the complex financial landscape.

\begin{table}[ht]
    \caption{\textnormal{Performance comparison with Multi-scale on S\&P100, CSI300 and CSI500.Multiscale number K={2,3,4}.}}
    \scriptsize
    \setlength{\arrayrulewidth}{0.7pt}
    \renewcommand{\arraystretch}{1.5}
    \centering
    \setlength{\tabcolsep}{4pt}
    \begin{tabular}{>{\hspace{10pt}}c c c c c }
    \hline
    \multicolumn{1}{>{\hspace{8pt}}c}{\cellcolor[rgb]{0.867, 0.875, 0.91}Dataset} & 
    \multicolumn{1}{c}{\cellcolor[rgb]{0.867, 0.875, 0.91}Metrics} &
    \multicolumn{1}{c}{\cellcolor[rgb]{0.867, 0.875, 0.91}K=2} & 
    \multicolumn{1}{c}{\cellcolor[rgb]{0.867, 0.875, 0.91}K=3} & 
    \multicolumn{1}{c}{\cellcolor[rgb]{0.867, 0.875, 0.91}K=4} \\
    
    \hline
    \rowcolor[rgb]{.949, .949, .949}
    \multirow{2}{*}[0.5em]{\shortstack{S\&P100}} & ACC & 52.60& \textbf{53.35} & 52.18\\
    \rowcolor[rgb]{.949, .949, .949}
     & MCC & 0.0290 & \textbf{0.0612} & 0.0231\\
    \hline
    
    \multirow{2}{*}[0.5em]{CSI300} & ACC & 50.21 & \textbf{53.21} & 51.47\\
     & MCC & 0.0352 & \textbf{0.0567} & 0.0404\\
    \hline
    
    \rowcolor[rgb]{.949, .949, .949}
    \multirow{2}{*}[0.5em]{CSI500} & ACC & 52.14 & \textbf{53.46} & 53.26 \\
    \rowcolor[rgb]{.949, .949, .949}
     & MCC &0.0286 & \textbf{0.0651} & 0.0392 \\
    \hline
    \end{tabular}
    \label{tab:my_label}
\end{table}

\subsection{Ablation Study}

In order to further analyze the effectiveness of our proposed module, we have divided it into three sub-models, as described below. Figure 3 shows the performance results of the ablation experiment under the binary classification of stock trend prediction.

\begin{itemize}
    \item \textbf{w/o features-SR:} To verify the effectiveness of features SR in capturing the internal relationships of stocks and the relationships between stocks in the entire market, we removed the features SR section based on MS-HGFN.
    
    \item \textbf{Replaced with concatenations' MSF:} To evaluate the effectiveness of multi-scale fusion in integrating feature information from different time scales. We will change the fusion part to direct concatenation.
    \item \textbf{Replaced with LSTM:} To verify the effectiveness of Transformer in capturing temporal dependencies, we replaced Transformer with LSTM based on MS-HGFN.
    
\end{itemize}

As shown in Fig. 3, the excellent performance of MS-HGFN validates the positive impact of each module. Traditional models that overlook the features of individual stocks pose challenges to prediction accuracy. The features-SR module analyzes the internal characteristics of stocks to reveal complex interactions. The MSF module dynamically fuses multi-scale features through gating mechanisms to filter noise. The replacement of LSTM with Transformer confirms that the latter demonstrates superiority in capturing long-term dependencies via self-attention, which is suitable for the non-local features of stock trends.

\begin{figure}[ht]
    \centering
    \includegraphics[width=\textwidth]{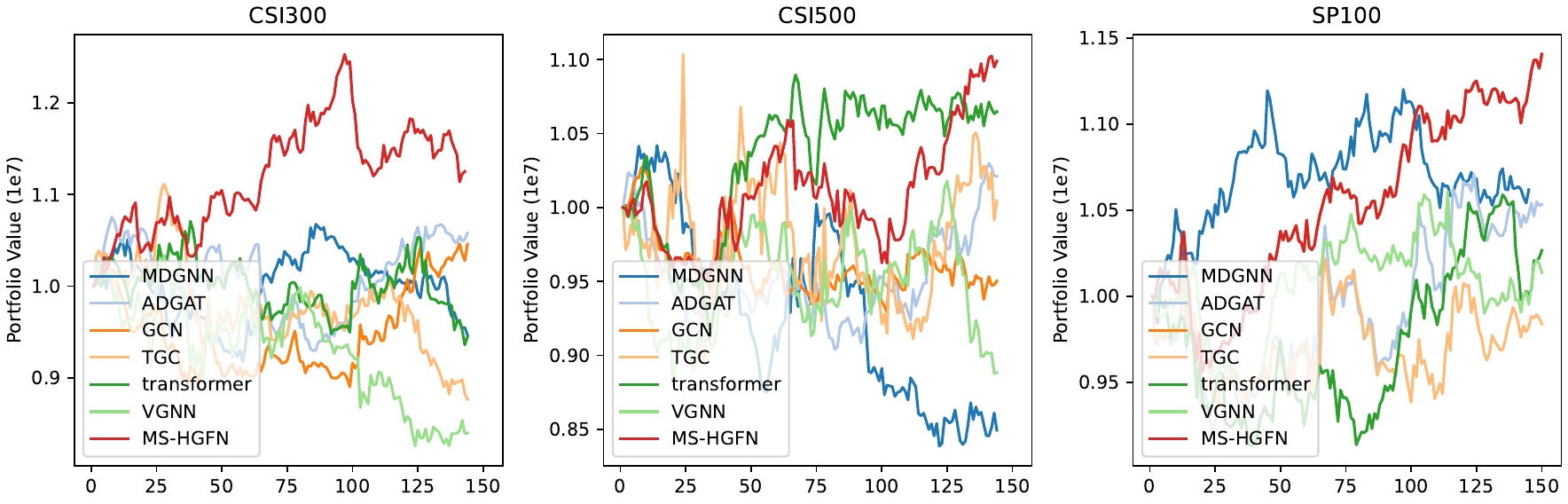}
    \caption{Performance diagram of backtesting for all methods.}
    \label{fig:sample-image}
\end{figure}

\subsection{Research on backtesting returns}

We assessed portfolio value changes using datasets CSI300, CSI500, and S\&P100, focusing on top-five stock returns. Starting with 10 million, the "MS-HGFN" model consistently outperformed others in stability and growth. In the CSI300 dataset, its value peaked at 12 million by the 100th day, outperforming models like MDGNN and ADGAT. For the CSI500, it grew steadily to 11 million by day 120, surpassing models such as TGC and transformer. In the S\&P100 dataset, "MS-HGFN" achieved a value of 11.5 million, showcasing superior market adaptability. Other models, like GCN and VGNN, fell below the initial investment. Overall, "MS-HGFN" demonstrated excellent performance in peak value and stability, proving effective for top-tier investment portfolios.

\subsection{Comparison of Multi-scale feature Extraction}

In our analysis, we assessed the impact of different scale settings for the MS-HGFN model using datasets S\&P100, CSI300, and CSI500, focusing on parameter \( K = \{2, 3, 4\} \). The results showed that \( K = 3 \) generally provided the best performance across all datasets. Specifically, the model achieved a 53.35\% accuracy and 0.0590 MCC on the S\&P100 dataset. For CSI300, it recorded an ACC of 53.21\% and MCC of 0.0567, and on CSI500, it led with an ACC of 53.46\% and MCC of 0.0651. Although \( K = 4 \) maintained high accuracy in CSI300, its MCC was slightly lower than with \( K = 3 \).

Overall, setting \( K = 3 \) demonstrated superior adaptability to market dynamics, particularly in the Chinese A-share market. Variations with \( K = 2 \) and \( K = 4 \) exhibited more volatility, underscoring the importance of aligning multi-scale window complexity with specific market characteristics. This is due to the short-term and long-term volatility of the stock market; using downsampling can better guide fine-grained capture of future trend features in time series when making trend classification predictions. However, too many down-sampling layers can introduce more coarse-grained features, reduce the ability to identify cross-scale shared features, and affect prediction performance.

\section{Conclusion}

Our research introduces the MS-HGFN framework to improve stock trend forecasting by overcoming the constraints in current multi-scale GNN techniques. The approach skillfully merges multi-scale sampling, dynamic hierarchical relationship modeling, and multi-scale feature fusion, capturing multifaceted dependencies between stocks. Experiments on U.S. and Chinese stock datasets show that MS-HGFN consistently surpasses most SOTA methods in ACC and MCC metrics, confirming its enhanced prediction accuracy.

\section{Acknowledgments}

This study is supported by Northeastern University, Shenyang, China (02110022124005 and 0211007342300).

\bibliographystyle{unsrt} 
\bibliography{ref} 
\end{document}